\documentclass[runningheads]{llncs}
\usepackage{subcaption} 
\usepackage{float}
\usepackage{pifont} 
\usepackage[T1]{fontenc}
\usepackage{booktabs}
\usepackage{tabularx}   
\usepackage{makecell}   
\newcolumntype{Y}{>{\centering\arraybackslash}X} 
\usepackage{amsmath}
\usepackage{amssymb}   
\usepackage{mathtools} 
\usepackage{multirow}
\usepackage{array}
\usepackage{graphicx}
\usepackage[misc]{ifsym}

\begin{document}
\title{Improving Multi-turn Dialogue Consistency with Self-Recall Thinking}
%
%

\author{Renning Pang\and
Tian Lan \and
Leyuan Liu \Letter \and  
Xiaoming Huang \and
Piao Tong \and
Xiaosong Zhang}
\authorrunning{R. Pang et al.}
%
\institute{
University of Electronic Science and Technology of China\\
\email{prn@std.uestc.edu.cn}, 
\email{lantian1029@uestc.edu.cn}, 
\email{leyuanliu@uestc.edu.cn},  
\email{piaot@std.uestc.edu.cn}, 
\email{johnsonzxs@uestc.edu.cn}
}
\maketitle              
\begin{abstract}
    Large language model (LLM) based multi-turn dialogue systems often struggle to track dependencies across non-adjacent turns, undermining both consistency and scalability. As conversations lengthen, essential information becomes sparse and is buried in irrelevant context, while processing the entire dialogue history incurs severe efficiency bottlenecks. Existing solutions either rely on high latency external memory or lose fine-grained details through iterative summarization. In this paper, we propose Self-Recall Thinking (SRT), a framework designed to address long-range contextual dependency and sparse informative signals in multi-turn dialogue. SRT identifies helpful historical turns and uses them to generate contextually appropriate responses, enabling the model to selectively recall and reason over context during inference. This process yields an endogenous reasoning process that integrates interpretable recall steps without external modules. SRT incorporates: (1) Dependency Construction: Generating and converting it into self-recall chains; (2)Capability Initialization: Training to enable reasoning chains with recall tokens capability; (3)Reasoning Improvement: Refining accuracy via verifiable rewards to optimize recall and reasoning for correct answers. Experiments on multiple datasets demonstrate that SRT improves F1 score by 4.7\% and reduces end-to-end latency by 14.7\% over prior methods, achieving a balance between reasoning latency and accuracy, and outperforming state-of-the-art baselines.

\end{abstract}
\section{Introduction}
Multi-turn dialogue is central to real-world assistants in customer service and instant messaging, where sustained, context-aware exchanges drive user satisfaction and operational efficiency \cite{ibm_generative_ai}. However, this requirement creates a fundamental tension \cite{zhang2024business} as conversations lengthen, the volume of historical context grows, yet the specific knowledge needed for a coherent response becomes increasingly sparse. Simply processing this entire, ever-growing history is not a viable solution, as it leads to significant performance bottlenecks, including attention dilution and prohibitive computational costs \cite{zhang2023mind}. This challenge forces reliance on imperfect strategies and highlights a critical research gap regarding the need for a mechanism that can leverage long-term dialogue history with both precision and efficiency \cite{aws_ai_agents}.

\begin{figure}
\centering
\includegraphics[width=\textwidth]{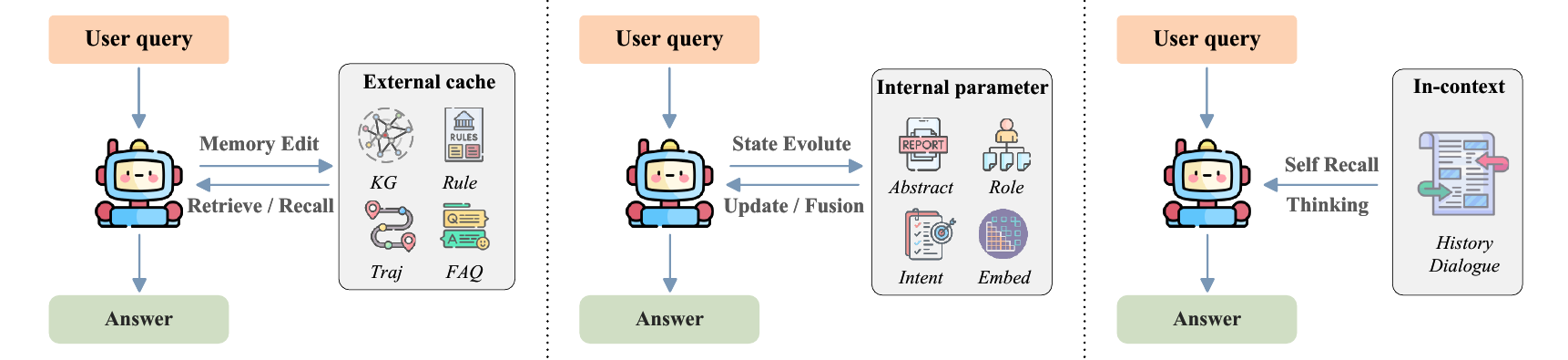}
\caption{Comparison of SRT with other multi-turn dialogue generation approaches} \label{fig1}
\end{figure}

As shown in Fig. \ref{fig1}, the utilization of history as the engine of such multi-turn dialogues is typically manifested in two dominant paradigms. One major line of work focuses on structuring memory externally, using components like graph databases or explicit episodic logs to organize dialogue history. However, this approach introduces significant system latency, complex workflow design, and high costs for multi-agent deployment \cite{chen2025compress}. Another popular approach involves learning an adaptive summary policy, often using alignment tuning to decide state evolution. These methods offer flexibility, but they typically operate by updating a condensed or abstractive memory state, where crucial latent details are progressively lost during compressive update \cite{tan2025prospect,pan2025memory}. A common thread across these approaches is their reliance on external architectures, complex preprocessing, or non-trivial learning processes, which introduce latency and loss of detailed knowledge.

To bridge these limitations, we propose the Self-Recall Thinking (SRT) mechanism and design an adaptive endogenous explicit recall strategy to balance precision and efficiency, enabling the model to mine and recall historical knowledge rather than passively retrieve. Instead of summarizing the entire dialogue context at a coarse level, we argue that during the reasoning process, models should proactively analyze and recall only the key historical turns directly relevant to the current query. SRT represents a transition from externally-oriented refinement to endogenous recall of fine-grained dependent content. During reasoning, the model can analyze the current query to recall relevant historical turns, generating a predefined signal to replay in thinking on-demand. Specifically, SRT follows an $analyze$ → $recall$ → $cite$ → $reason$ → $answer$ cycle within each turn. For example, in a customer service dialogue, when asked whether a membership discount applies to a specific booking at the Uptown branch, SRT opens $\texttt{<HIS>}$ tag to cite the exact turns that contain the discount rule that applies above one hundred fifty dollars, the service price of one hundred eighty dollars, and the Uptown location: {\ttfamily
<HIS> Q1: I need to check membership status. I'm under Apex Tier. A5: The Chakra Balancing service is \$180. A6: Lena only \allowbreak
offers that service at our Uptown branch.</HIS>}, then continues thinking to conclude that the discount applies. We also propose a novel adaptive endogenous explicit recall strategy SRT-P, which can be directly applied to dialogue agent invocation scenarios. Empirically, the experimental results confirm that SRT effectively recalls and exploits dialogue history for multi-turn reasoning, surpassing prior state-of-the-art systems with an average F1 increase of 3.4\% and an average latency reduction of 13.1\% across QA benchmarks, each benchmark gains range from 1.1\% to 4.7\% in F1 and from 11.1\% to 14.7\% in end-to-end latency. Additionally, the pluggable strategy abstracted from SRT mechanism can be applied to closed-source commercial models to gain revenue.
The key contributions of this work are summarized as follows:
\begin{itemize}
  \item We introduce SRT, a novel reasoning strategy for multi-turn dialogues. SRT enables the model to proactively recall and selectively attend to dependent content from the dialogue history, thereby enhancing reasoning accuracy with fine-grained details.
 \item We constructed instruction fine-tuning datasets and trained the model via fine tuning alignment to equip the model with self-recall capabilities. Furthermore, we designed a verifiable reward mechanism to guide the optimization of its recall behavior, ensuring that the model effectively leverages historical context.
 \item We conducted extensive experiments on several multi-turn dialogue benchmarks. The results demonstrate that our proposed model significantly outperforms existing baseline approaches on key downstream tasks, achieving a superior trade-off between computational cost and reasoning accuracy.
\end{itemize}

\section{Related Work}
\subsection{External Cache Retrieval}
Early retrieval-augmented systems for dialogue agents stored raw or summarized conversation history and relied primarily on vector similarity search \cite{chan2024rq}. Recent efforts have introduced varying degrees of structural organization into memory. Knowledge graph based methods \cite{xu2024generate,jiang2024kg,dammu2025dynamic} build explicit relational structures among entities or events, often augmented with temporal edges to support multi-hop reasoning across sessions in practical deployments. A-MEM \cite{xu2025mem} adopts agentic organization, dynamically indexing and linking memory units as new evidence arrives. DH-RAG \cite{zhang2025dh} employs a hierarchical design that clusters content into multi-level schemas to route queries across abstraction levels. ComoRAG \cite{wang2025comorag} triggers targeted probes into memory, integrates retrieved evidence into a working state, and resumes inference. Extensions such as MemoRAG \cite{qian2025memorag} relax assumptions about query clarity and knowledge structure, improving long-context understanding at the cost of increased latency and controller complexity. EM-LLM \cite{fountas2024human} segments conversational streams into coherent episodes and retrieves based on both semantic relevance and temporal contiguity, enabling human-like memory access patterns. SGMem \cite{wu2025sgmem} extends this idea to dialogue by linking turns, rounds, and sessions, while mixing raw history with generated summaries to supply coherent multi-granular context. TiM \cite{liu2023think} stores and reuses thoughts rather than raw text, pairing pre-answer recall with post-answer updates, and employs locality-sensitive hashing for scalable retrieval.

\subsection{Latent State Evolution}
A complementary research direction internalizes memory within the LLMs, learning to maintain and update a compact latent state that co-evolves with the reasoning process. This paradigm can be organized into several interrelated strands. Memory-R1 \cite{yan2025memory} trains a Memory Manager and an Answer Agent using outcome driven reinforcement learning, turning memory management into a learned policy rather than a fixed pipeline. MemAgent \cite{yu2025memagent} learns an overwrite policy via RL to enable efficient long-text processing with constant size memory. Memento \cite{zhou2025memento} further supports continual adaptation through online RL schemes that update case-selection policies and rewrite episodic memory. ReasoningBank \cite{ouyang2025reasoningbank} complements these by distilling reusable reasoning strategies into a memory bank that is retrieved and scaled at test time. Another strand targets learning the latent memory state itself. MEM1 \cite{zhou2025mem1} maintains a constant size internal state that is jointly updated for memory and reasoning each turn, discarding redundancies while integrating new observations for better long-horizon generalization. Mem-$\alpha$ \cite{wang2025mem} employs RL to learn not only what information to extract and how to structure it across core, episodic, semantic components, and when to update it. MELODI \cite{husom2024price} introduces a hierarchical compression mechanism operating on short-term and long-term memory compresses and aggregates information at a mid-layer. Furthermore, MemGen \cite{zhang2025memgen} incorporates a memory trigger to decide when to recall and a memory weaver that synthesizes latent tokens within the reasoning process, yielding emergent planning and procedural memories.

\begin{figure}[t]
\centering
\includegraphics[width=\textwidth]{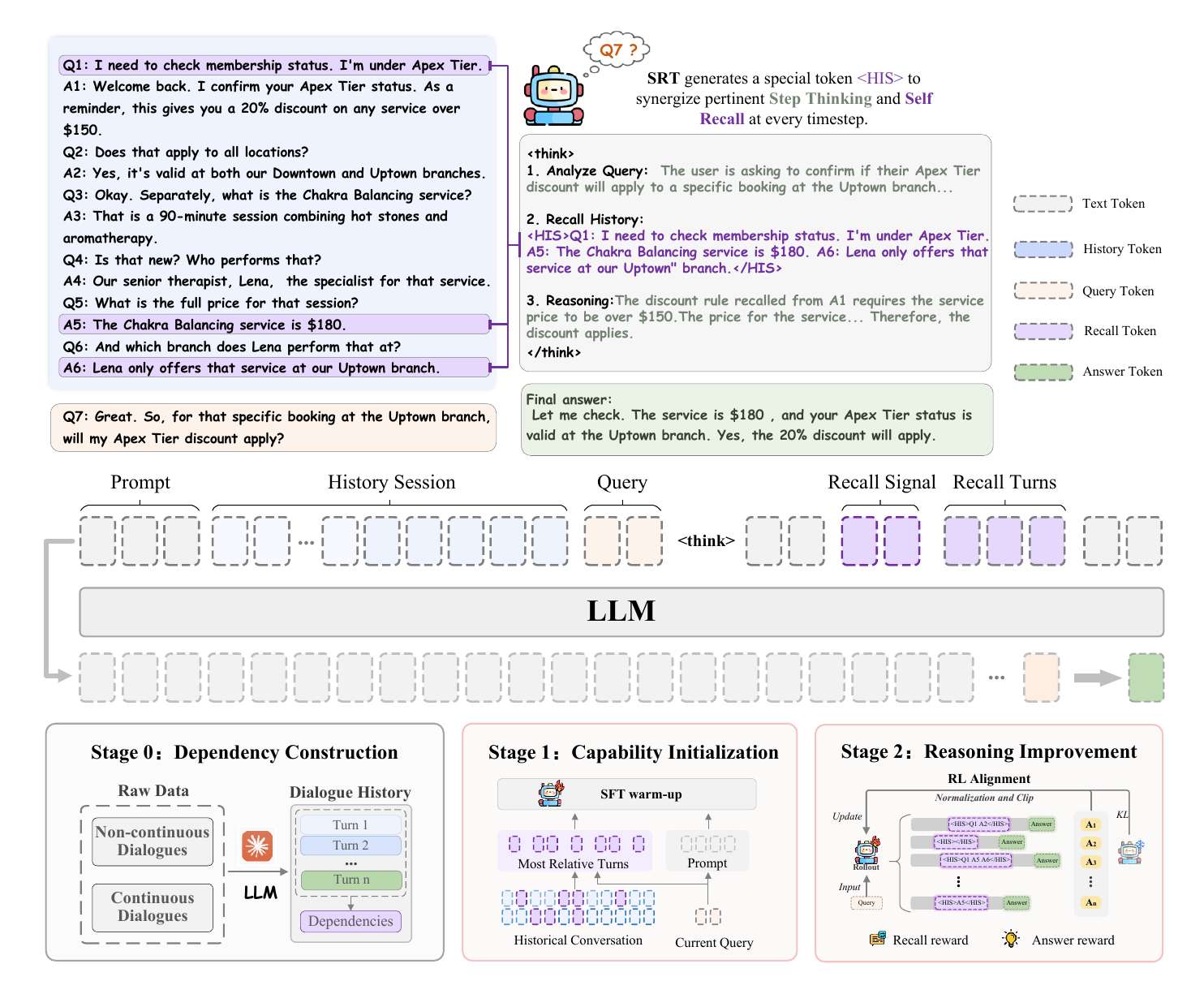}
\caption{SRT Framework. In Stage 0, we construct historical dependency structures from continuous/discontinuous dialogues. In Stage 1, we train the model to learn self-recall capability. In Stage 2, we refine the model's self-recall reasoning capability through RL using verifiable rewards.} \label{fig2}
\end{figure}

\section{Method}
\label{method}
As shown in Fig. \ref{fig2}, our approach is realized through a three-stage framework. Stage 0: Dependency Construction, which leverages Claude 3.7 Sonnet to convert raw data into a structured dialogue history, complete with corresponding dependencies. Stage 1: Capability Initialization, which involves Supervised Fine-Tuning to train the model to generate $\texttt{<HIS>}$ tag during the thinking process encapsulating recall content. Stage 2: Reasoning Improvement, which utilizes Group Relative Policy Optimization (GRPO) \cite{shao2024deepseekmath} to treat the recalling and answering in each context-dependent turn as a joint optimization objective, using a verifiable outcome reward. 

\subsection{Preliminaries}
We model self-recall as an internal pointer policy $\pi_\theta$ that operates over the dialogue history $C_t = \{u_1, ..., u_{t-1}\}$. At turn $t$, given the query $Q_t$, the model generates a CoT reasoning. During this process, the model can take $N$ recall actions. Each action, $o_k$ (for $k=1...N$), consists of emitting a $\texttt{<HIS>}$ tag that verbatim-copies a selected historical utterance $u_{i_k}$ from $C_t$. The set of recalled information $H_k$ is built recursively: $H_k = H_{k-1} \cup \{o_k\}$, with $H_0 = \emptyset$. The model implicitly terminates this recall phase (determining the final count $N$) by proceeding to generate the answer $A_t$, which is conditioned on the final recall set $H_N$. This mechanism treats the dialogue state update as a deterministic transition. The joint probability of generating $N$ recall actions ($o_{1:N}$) and the final answer $A_t$ is factorized as:

\begin{equation}
P(A_t, o_{1:N} | Q_t, C_t) = \left[ \prod_{k=1}^{N} \pi_\theta(o_k | Q_t, C_t, H_{k-1}) \right] \cdot \pi_\theta(A_t | Q_t, C_t, H_N)
\end{equation}

This formulation models the probability of each recall action $o_k$ as dependent on the query, context, and all previously recalled turns $H_{k-1}$. The final answer $A_t$ is then conditioned on the complete set of recalled evidence $H_N$.

\subsection{Dialogue Dependency Structure Construction}
\label{DATAS}

To curate a dataset focused on multi-turn context recall, we first implement a multi-stage dialogue filtering pipeline \cite{guo2025deepseek}. We begin by performing redundancy reduction and deduplication, applying clustering on intent features and turn-level feature hashing to merge or remove semantically identical dialogues. We then retain only dialogues with turn counts ranging from 8 to 32. The core of our filtering is a dependency-aware selection process. A dialogue is selected for annotation only if it contains at least one turn $Q_t$ (the current query at turn $t$) and a corresponding historical utterance $u_i$ (at turn $i$) that satisfy a dual-dimensional scoring function:

\begin{equation}
\text{Score}(u_i, Q_t) = 0.6 \cdot \text{Sim}_{\text{sem}}(u_i, Q_t) + 0.4 \cdot e^{-0.15 \cdot |t-i|} > 0.6
\end{equation}

 $\text{Sim}_{\text{sem}}$ denotes semantic similarity, ensuring the historical turn $u_i$ is relevant. The second term is an exponential decay factor as 0.15 is an empirically optimal value that reduces the weight of historical utterances from 10 rounds prior to 0.22. This selection process ensures that the retained dialogues contain quantifiable, non-adjacent dependencies, forcing the model to retrieve information from distant history and directly aligning with SRT's recall-oriented goal.

To generate reliable supervision signals for recall and reasoning, we design a hybrid annotation pipeline. For each turn $(Q_t, C_t)$ in the filtered dialogues, we use Claude 3.7 Sonnet \cite{lior2025reliableeval} as a teacher model, guided by recall-targeted signals. The model is prompted endogenously to produce the minimal recall set $H_t^*$ as the smallest set of historical utterances required to answer $Q_t$, formatted as \texttt{Turn X: 'content'} to enable turn-level recall. Then we get the step-by-step reasoning from recall to concise answer, which is strictly grounded in the content of $H_t^*$, designed to facilitate alignment with reward designs. Moreover, we apply a data-driven verification filter to the generated triplets $(H_t^*, Z_t, A_t^*)$. We ensure logical consistency between the thinking $Z_t$ and the recall set $H_t^*$, and verify strict answer--evidence alignment by confirming no hallucination via string matching between $A_t^*$ and $H_t^*$. 

\subsection{Supervised Fine-Tuning for Capability Warm-up}

Despite the availability of explicit recall annotations, directly training language models on raw multi-turn dialogues often fails to induce stable CoT behaviors, as models tend to overfit surface patterns or generate ungrounded rationales when processing lengthy, heterogeneous contexts. To address this, we develop a two-stage training approach that first bootstraps recall capabilities using dependency pruned histories, then transfers these capabilities to full contextual reasoning.

\subsubsection{Bootstrapping with Dependency-Pruned Histories}

For each dialogue turn $t$, we construct simplified training instances by pruning the complete history $C_t$ to a compact subset $\tilde{C}_t$ using the dependency scoring mechanism from Section~2. From the teacher-generated annotations, we preserve the minimal recall set $H_t^* \subset \tilde{C}_t$, the structured reasoning chain $Z_t = \{r_1, \dots, r_M\}$ with explicit \texttt{<HIS>} tagging, and the evidence-grounded answer $A_t^*$. This pruning strategy reduces contextual distractions while maintaining logical dependencies, providing clean training signals for learning recall timing, target selection, and information utilization.

\subsubsection{Reasoning Transfer to Full Contexts}

The distilled recall reasonings subsequently guide rationale generation on complete dialogue histories. Given a prompt template $\mathcal{P}(\cdot)$ and target outputs $(Z_t, A_t^*)$, we optimize the negative log-likelihood objective:

\begin{equation}
\mathcal{L}_{\text{SFT}} = -\sum_{i=1}^{M} \log \pi_\theta \left( r_i \mid \mathcal{P}(Q_t, C_t), r_{<i} \right) - \sum_{j=1}^{L} \log \pi_\theta \left( a_j \mid \mathcal{P}(Q_t, C_t), Z_t, a_{<j} \right)
\end{equation}

This staged training approach effectively activates the model's inherent reasoning capabilities for the self-recall task, establishing a robust foundation for subsequent alignment enhancement.

\subsection{Recall-Reasoning Alignment}
\subsubsection{Verifiable Composite Reward Design.} To refine the model's reasoning capability and prevent reward hacking behaviors, we design a composite reward function $R(\tau)$ that comprehensively evaluates the entire generated reasoning $\tau = (Z_t, A_t)$. This function is fully verifiable, as all components are computed by comparing the model's outputs against explicit rule-based criteria and ground-truth annotations, eliminating dependence on learned reward models that may introduce bias or instability.

Given a reasoning $\tau$ comprising the reasoning chain $Z_t$ and final answer $A_t$, we first extract the predicted recall set $\hat{H}_t$ by aligning each \texttt{<HIS>} tag in $Z_t$ to its corresponding historical utterance $u_i \in C_t$ through maximal normalized overlap matching. The comprehensive reward function is then defined as a weighted combination of three core components:

\begin{equation}
R(Z_t, A_t) = \text{format}(Z_t) + \text{recall}(\hat{H}_t, H_t^*) + \text{answer}(A_t, A_t^*)
\end{equation}

The format reward ensures the structural integrity of the recall mechanism and prevents malformed outputs that would compromise auditability. It is defined as a binary function that outputs 1 only if the reasoning trace $Z_t$ demonstrates complete syntactic validity, meaning all \texttt{<HIS>} tags are properly formatted, correctly paired, and parseable within established reasoning steps:

\begin{equation}
R_{\text{format}}(Z_t) = \mathbb{I}(\text{is\_valid}(Z_t))
\end{equation}

The Recall Reward ($R_{recall}$) provides a fine-grained evaluation of recall accuracy, creating balanced incentives for the model to identify all necessary dependent turns while avoiding extraneous recalls. It operates within the range $[-1.25, 1.25]$ and is calculated as a scaled Jaccard Index (Intersection over Union):

\begin{equation}
R_{recall}(\hat{H}_t, H_t^* \rvert) = 2.5 \cdot \left( \frac{\lvert \hat{H}_t \cap H_t^* \rvert}{\lvert \hat{H}_t \cup H_t^* \rvert} \right) - 1.25
\end{equation}
where $H_t^*$ represents the ground-truth recall set, and $\hat{H}_t$ denotes the predicted set. This formulation inherently penalizes both Missing Recall (which increases the union $\lvert \hat{H}_t \cup H_t^* \rvert$ via $H_t^*$) and Over-Recall, which also increases the union via $\hat{H}_t$. By rewarding the maximization of the intersection relative to the union, this design promotes both completeness and precision in the recall process.

The answer reward measures semantic quality of the final output, ensuring the model is rewarded for producing answers that are semantically equivalent to the reference even when using different phrasing or expression. It operates within the range $[-1, 1]$ and is computed as:

\begin{equation}
R_{\text{answer}}(A_t, A_t^*) = \text{CosineSimilarity}(\vec{v}_{A_t}, \vec{v}_{A_t^*})
\end{equation}

This reward provides a robust similarity measure beyond surface-level token matching, which ensures the model is optimized not only for the quality of the final output, but also for the fidelity and accuracy of the internal reasoning process that produces it, thereby promoting both performance and interpretability in self-recall thinking while maintaining training stability through verifiable, rule-based signal computation.

\subsubsection{Policy Optimization with RL.} We adopt GRPO for policy optimization due to its memory efficiency and elimination of value network training requirements. This approach simplifies alignment by leveraging group-relative advantages to normalize learning signals across reasonings.

For each prompt $(Q_t, C_t)$, we sample $K$ reasonings $\{\tau^{(k)}\}_{k=1}^K$ from the behavior policy $\pi_{\theta_{\text{old}}}$ and compute their respective rewards $R^{(k)} = R(\tau^{(k)})$. We then calculate group-relative advantages to reduce variance across reasonings:

\begin{equation}
\tilde{A}^{(k)}=\frac{R^{(k)}-\bar{R}}{\sigma_R+\epsilon} \quad r_{i, t}(\theta)=\frac{\pi_\theta\left(o_{i, t} \mid q, o_{i,<t}\right)}{\pi_{\theta_{\text {old }}}\left(o_{i, t} \mid q, o_{i,<t}\right)}
\end{equation}

where $\bar{R}$ represents the group mean reward, $\sigma_R$ denotes the group standard deviation, and $\epsilon = 10^{-8}$ prevents division by zero. The complete GRPO optimization objective is formalized as:

\begin{equation}
\begin{aligned}
\mathcal{J}_{\mathrm{GRPO}}(\theta)
&=\mathbb{E}_{(q,a)\sim\mathcal{D},\{o_i\}_{i=1}^{G}\sim\pi_{\theta_{\mathrm{old}}}(\cdot\mid q)}
\left[\frac{1}{G}\sum_{i=1}^{G}\frac{1}{|o_i|}\sum_{t=1}^{|o_i|}\bigl(
\min\bigl(r_{i,t}(\theta)\hat{A}_{i,t},\right.\\[4pt]
&\hspace{6em}\left.\operatorname{clip}\bigl(r_{i,t}(\theta),1-\varepsilon,1+\varepsilon\bigr)\hat{A}_{i,t}\bigr)
-\beta D_{\mathrm{KL}}(\pi_\theta\|\pi_{\mathrm{ref}})
\bigr)\right]
\end{aligned}
\end{equation}

where $\pi_{\theta}$ is optimized by maximizing a clipped surrogate objective, augmented with a token-level KL divergence term that anchors to the supervised fine-tuned policy $\pi_{\text{ref}} = \pi_{\text{SFT}}$ to prevent catastrophic drift from Stage 1 capabilities.

\section{Experiment}
\subsection{Implementation Details}
We adopt Qwen2.5-7B \cite{team2024qwen2} as the backbone model to ensure a strong and consistent baseline. Consistent with the two stage training framework outlined in Section \ref{method}, we implement SFT and RL alignment for model training, with all experiments conducted on 8 NVIDIA A100 (80 GB) GPUs to support efficient training of the 7B-scale model. For Stage 1, the model is trained with a learning rate of $5\times10^{-6}$. For Stage 2, we reduce the learning rate to $3\times10^{-7}$ to stabilize RL updates, leveraging GRPO algorithm for RL and incorporating a KL divergence regularization term. We employ the AdamW optimizer with a constant learning rate schedule supplemented by linear warm-up. We use a rollout size of 8, and set the ratio of the sample batch size to the backpropagation batch size to 4.

\subsection{Datasets}
To evaluate SRT, we use a diverse set of benchmarks. First, we created SRQA, a unified dataset of 5,000 samples (8-32 turns) emphasizing long-range dependencies. It was curated from MG-ShopDial \cite{bernard2023mg}, MultiWOZ \cite{budzianowski2018multiwoz}, DailyDialog \cite{li2017dailydialog}, ReDial \cite{zhou2020towards} and LoCoMo \cite{maharana2024evaluating} using the processing method described in Section \ref{DATAS}.
Moreover, we include two standard open-domain QA benchmarks to assess general reasoning: SimpleQA \cite{wei2024measuring} contains 4,326 short, factual questions, where each question has only one clear answer. CoQA \cite{reddy2019coqa} is a  simulate real-life dialogue question answering dataset with over 127,000 question-answer pairs from 8,000 conversations.

\subsection{Baselines}
We compare our method against 6 representative baselines from three paradigms, built upon the Qwen2.5-7B foundation model for a fair comparison. We employed single-model inference in our tests rather than multi-agent workflow mechanisms. For the extra augmented paradigm, RQ-RAG \cite{chan2024rq} for ambiguous or complex queries, first performs query rewriting and decomposition to eliminate ambiguity before searching. QRMeM \cite{wang2024qrmem} maintains a dual-pool memory that fuses static text with structured graph information and navigates retrieval via reflective trial and error. LD-Agent \cite{li2025hello} employs separate long and short term memory banks with topic-aware retrieval and dynamic persona modeling. For the latent reasoning paradigm, Coconut \cite{hao2024training} feeds the LLM's final hidden states back as continuous thoughts into the input embedding space, enabling parallel exploration of reasoning paths. OPRO \cite{zhang2024revisiting} uses the language model itself as an optimizer, framing the search for an optimal reasoning instruction as a formal optimization task. SoftCoT \cite{xu2025softcot} employs a frozen lightweight assistant to generate soft thought tokens projected into the backbone model's representation.

\begin{table}[htbp]
\centering
\renewcommand{\arraystretch}{1}
\setlength{\tabcolsep}{6pt}
\caption{Main results on three benchmarks. Metrics are F1 and average end-to-end latency (s).}
\footnotesize
\label{tab:main_results_qa}
\begin{tabular*}{\linewidth}{@{\extracolsep{\fill}} llcc|cc|cc}
\hline
\multirow{2}{*}{Category} & \multirow{2}{*}{Method}
& \multicolumn{2}{c|}{SRQA}
& \multicolumn{2}{c|}{CoQA}
& \multicolumn{2}{c}{SimpleQA} \\
\cline{3-8}
& & F1 & Latency & F1 & Latency & F1 & Latency \\
\hline
\multirow{3}{*}{Extend Cache}
 & RQ\text{-}RAG   & 72.9 & 14.7 & 81.0 & 13.6 & 53.6 & 14.8 \\
 & QRMeM           & 70.6 & 12.5 & 75.2 & 12.1 & 50.3 & 12.9 \\
 & LD\text{-}Agent & 73.1 & 11.8 & 77.6 & 11.5 & 51.7 & 12.1 \\
\hline
\multirow{3}{*}{Latent Parameter}
 & OPRO            & 71.2 & 12.0 & 79.8 & 12.3 & 49.2 & 14.6 \\
 & Coconut         & 75.0 & 10.5 & 83.1 & 10.2 & 52.5 & 10.8 \\
 & SoftCoT         & 70.4 & 13.3 & 78.6 &  9.9 & 51.8 & 10.2 \\
\hline
Our & SRT
 & 78.4 & 9.1
 & 84.0 & 8.8
 & 56.1 & 8.7 \\
\hline
\end{tabular*}
\end{table}
\subsection{Overall Performance}
Table \ref{tab:main_results_qa} shows the main results across three question-answering benchmarks. Table 1 compares the F1 scores and end-to-end latency of SRT with baselines across three QA benchmarks. SRT achieves the optimal accuracy–efficiency trade-off universally: 56.1 F1/8.7 s on SimpleQA, 84.0 F1/8.8 s on CoQA, and 78.4 F1/9.1 s on SRQA. Across datasets, SRT outperforms baselines in both accuracy and efficiency: on SimpleQA, it surpasses RQ-RAG 53.6 F1 with 2.5\% improvement and Coconut 52.5 F1 with 3.6\% improvement, while being 2.1s–6.1s faster; on CoQA, it edges Coconut and RQ-RAG with the lowest latency; on SRQA, it outperforms Coconut by 3.4\% and LD-Agent by 5.3\%. Furthermore, with a latency of 9.1s, SRT is between 1.4s and 5.6s faster than the baselines, whose latencies range from 10.5s to 14.7s. We further compare SRT against two baseline paradigms. External Augmented Memory methods, which rely on external modules, introduce significant latency, with measured times ranging from 11.5s to 14.8s across the benchmarks. For instance, even the top performing method in this category, RQ-RAG exhibits a latency of at least 13.6s. While Latent-Memory methods generally offer lower latency than their external counterparts, they still fall short of the efficiency achieved by our approach. In contrast, SRT internalizes retrieval via $\texttt{<HIS>}$ tags, eliminating the need for external modules and providing token-level auditable grounding, which explains its simultaneous accuracy and speed advantages. SRT excels most in conversation-heavy and long-context scenarios: on SRQA, it outperforms LD-Agent with 5.3\% improvement and is 1.4s faster. On CoQA, SRT avoids over/missing-recall errors to maintain 84.0 F1 and 8.8s.

\subsection{Model design and optimization strategy}
We removed core components of the SRT from the integrated dialogue corpus. As shown in Table \ref{table2}, removing the reinforcement learning stage component resulted in the most significant performance degradation, a 7.2\% drop in historical recall and a 4.8\% decrease in answer accuracy. This confirms that the joint reward mechanism for recall and reasoning significantly influences the model's recall motivation. Disabling the CoT module also caused a noticeable performance drop, with accuracy decreasing by 3.4\%, confirming the effectiveness of explicit recall representations. The baseline using only prompts without fine-tuning performed worst, failing to effectively follow instructions within limited model capability. 
\begin{table}[ht]
\centering
\setlength{\tabcolsep}{6pt}
\renewcommand{\arraystretch}{1}
\caption{Ablation study results of SRT and its variants.}
\footnotesize
\label{table2}
\begin{tabular*}{0.7\linewidth}{@{\extracolsep{\fill}} lcc}
\hline
Variant & Recall (\%) & Acc. (\%) \\
\hline
SRT        & 92.5 & 78.9 \\
-- w/o RL  & 85.3 & 74.1 \\
-- w/o CoT & 89.1 & 76.0 \\
-- w/o SFT & 81.0 & 72.2 \\
\hline
\end{tabular*}
\end{table}

\begin{figure}[htbp]
\centering
\begin{subfigure}[t]{0.32\linewidth}
    \centering
    \includegraphics[width=\linewidth]{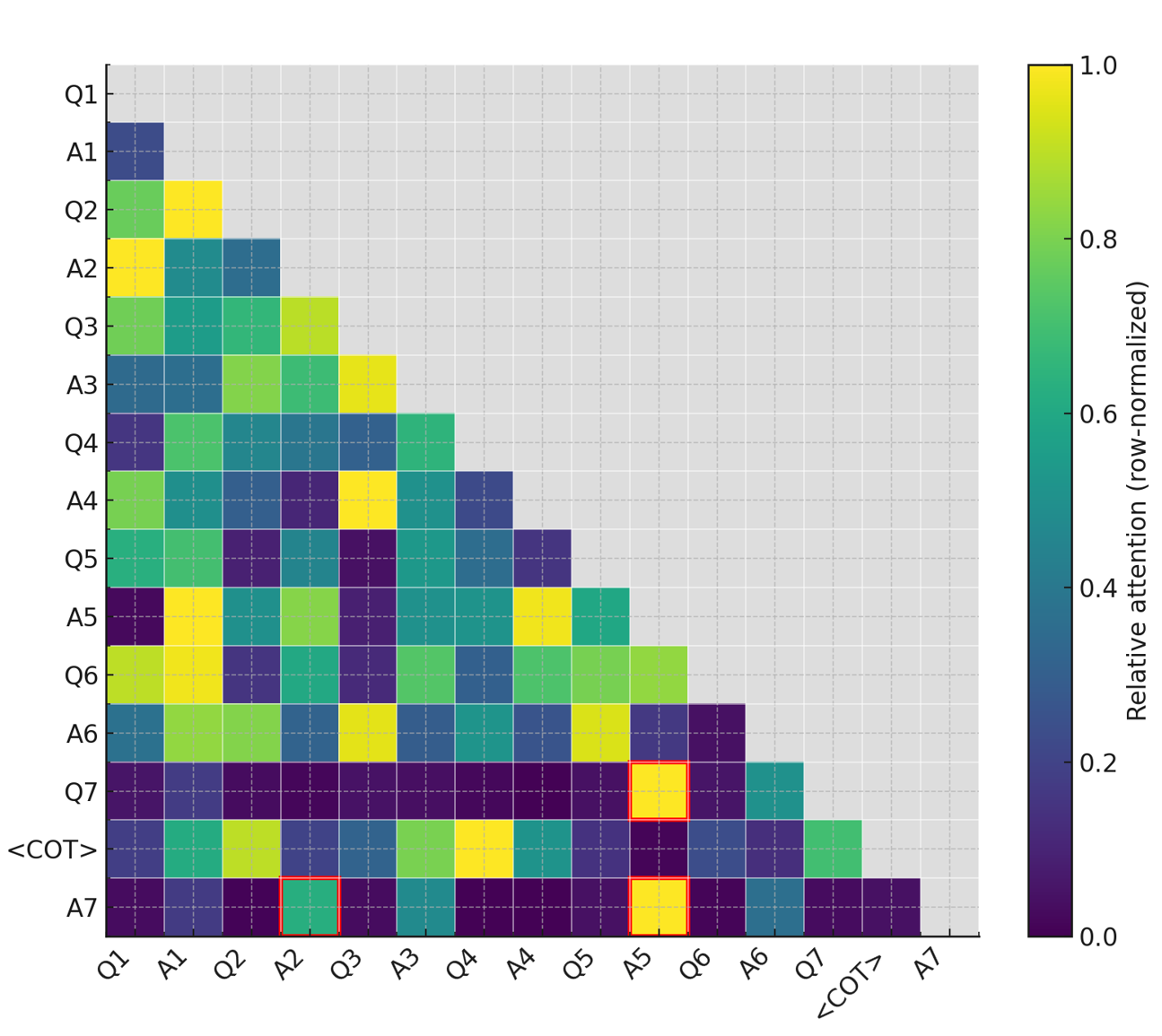}
    \caption{Non-recall guided attention map}
    \label{fig:attn_nonrecall}
\end{subfigure}
\hfill
\begin{subfigure}[t]{0.32\linewidth}
    \centering
    \includegraphics[width=\linewidth]{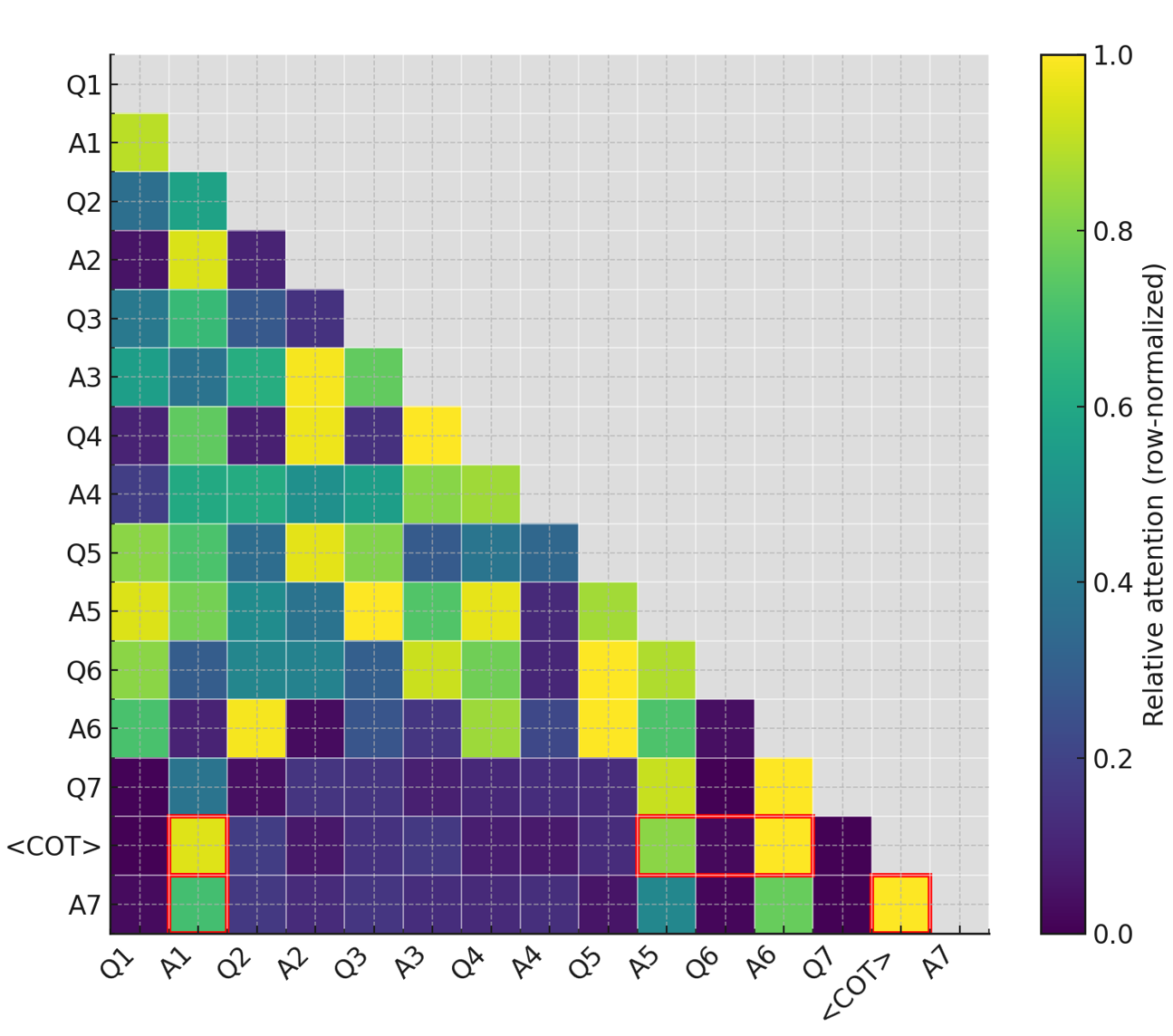}
    \caption{Self-Recall guided attention map}
    \label{fig:attn_recall}
\end{subfigure}
\hfill
\begin{subfigure}[t]{0.32\linewidth}
    \centering
    \includegraphics[width=\linewidth]{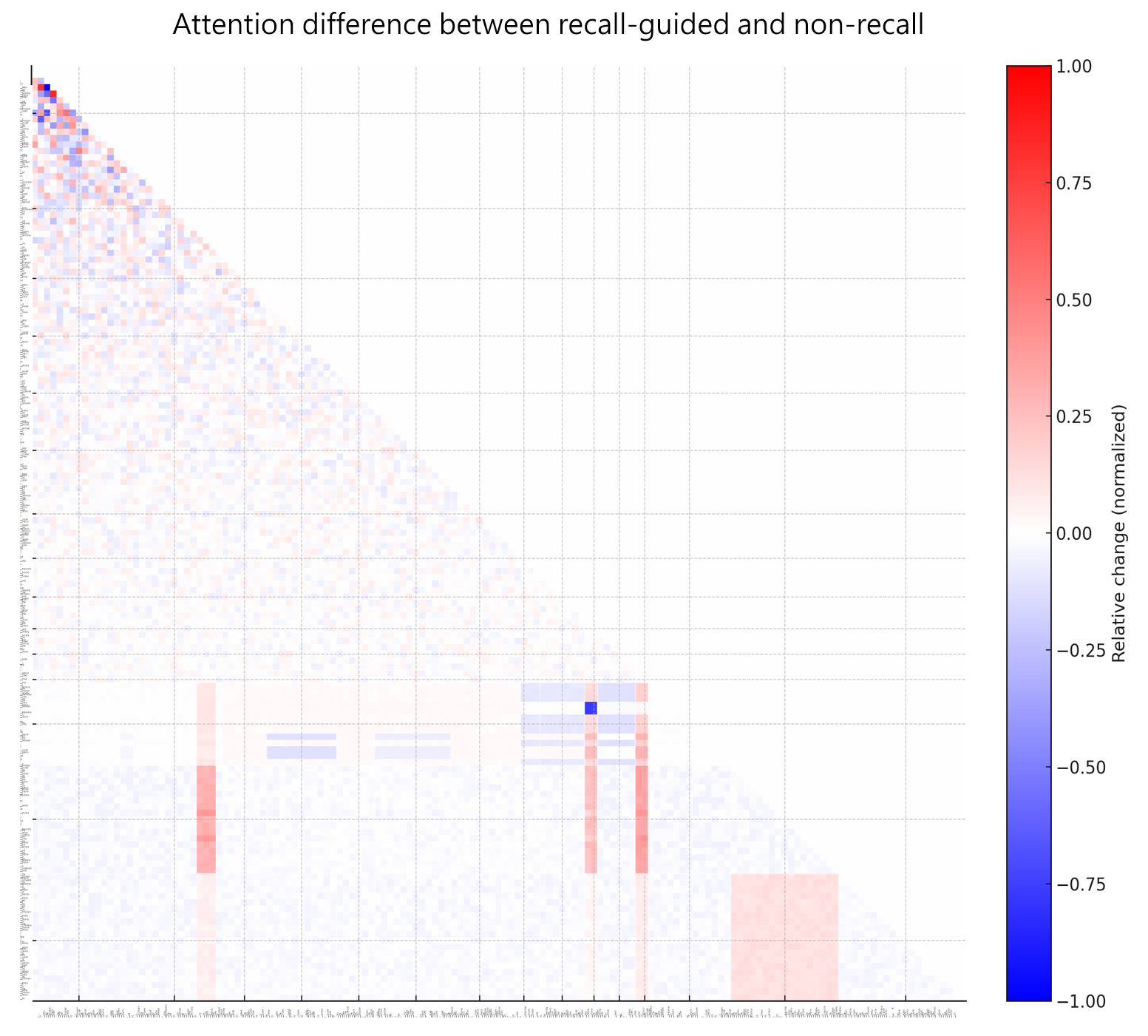}
    \caption{Attention residual Map}
    \label{Differences in attention allocation between recall-guided and non-recall-guided}
\end{subfigure}
\caption{Attention allocation visualization on the synthetic long-dialogue set.}
\label{atten}
\end{figure}
\subsection{Attention Allocation Comparison Analysis}
We analyze whether decoder attention concentrates on the turns retrieved by our self-recall mechanism during answer generation as shown in Fig. \ref{atten}. We compute attention on reasoning, the proportion of attention mass from answer tokens to the retrieved tokens, averaged over layers and normalized per row. We compare two settings: Non-Recall Fig. \ref{atten}(\subref{fig:attn_nonrecall}) and Self-Recall guided Fig. \ref{atten}(\subref{fig:attn_recall}). As dialogue length increases, the attention scores assigned to any single historical turn inevitably decrease. Critical information from early turns becomes lost in the middle, receiving negligible attention.

To better distinguish the gap, Fig. \ref{atten}(\subref{Differences in attention allocation between recall-guided and non-recall-guided}) visualizes differences in token-level attention distribution. SRT introduces an active, explicit recall mechanism that fundamentally alters the reasoning behavior. By generating $\texttt{<HIS>}$ tag, the model performs a discrete, high-salience action. It is no longer passively searching through a diluted context, which is actively pulling a critical piece of history out of the noise and placing it directly into its immediate working memory. This verbatim copying action forces the attention mechanism to focus on the recalled fact, bypassing the issues of dilution and interference for that piece of information.

\subsection{Incorporating failure analysis}
We analyze the proportion of each error type in the reasoning process among all bad cases. As shown in Table \ref{tab:error_distribution}, the contribution of Missing and Wrong-Recall errors increases with dialogue length, while the share of errors from Normal/No-Recall Needed, where the model fails despite correct recall, declines yet remains the single largest category. This trend can be interpreted through the lens of multi-turn customer service dialogues.

\begin{table}[H] 
\centering
\renewcommand{\arraystretch}{1}
\caption{Error type distribution over different conversation lengths.}
\footnotesize
\label{tab:error_distribution}
\begin{tabular*}{0.7\textwidth}{@{\extracolsep{\fill}} c|cccc}
\hline
$t$ (turns) & 
\begin{tabular}[c]{@{}c@{}}Missing\\ Recall\end{tabular} & 
\begin{tabular}[c]{@{}c@{}}Over\\ Recall\end{tabular} & 
\begin{tabular}[c]{@{}c@{}}Wrong\\ Recall\end{tabular} & 
\begin{tabular}[c]{@{}c@{}}Failure\\ Answer\end{tabular} \\ \hline
8  & 23.5\% & 5.3\% & 15.0\% & 56.1\% \\
16 & 28.2\% & 5.6\% & 16.7\% & 49.5\% \\
24 & 31.9\% & 5.7\% & 18.1\% & 44.3\% \\
32 & 35.9\% & 6.1\% & 19.2\% & 38.8\% \\ \hline
\end{tabular*}
\end{table}

The rise in Missing Recall occurs because crucial information such as order ID or account number is often provided in early turns and queried dozens of turns later, target clue becomes lost within para-service intervening turns like small talk or product inquiries etc. Using $\texttt{<HIS>}$ as a recall trigger mitigates this by performing relative turns recalled in CoT. The increase in wrong recall stems from the prevalence of same entities in long logs, where greater context and noise elevate the risk of false matches. In contrast, Over-Recall remains low and stable, slightly increasing as longer histories tempt the model to cite more segments.

\subsection{External evaluation of strategy effectiveness}
To verify the generalization of the self-recall mechanism beyond our fine-tuned SRT model, we distill it into a pluggable strategy SRT-P, that can be directly invoked via production APIs. We evaluate three closed-source large language models Claude 3.5 Sonnet \cite{anthropic2024claude}, DeepSeek-V3 \cite{liu2024deepseek}, and Qwen-Max \cite{yang2025qwen3} on a customer service style multi-turn workload. We report two key metrics answer accuracy and end-to-end latency across dialogue turn buckets \(k \in \{8,12,16,20,24,28,32\}\).

\begin{figure}[H]
\centering
\begin{subfigure}[t]{0.32\linewidth}
    \centering
    \includegraphics[width=\linewidth]{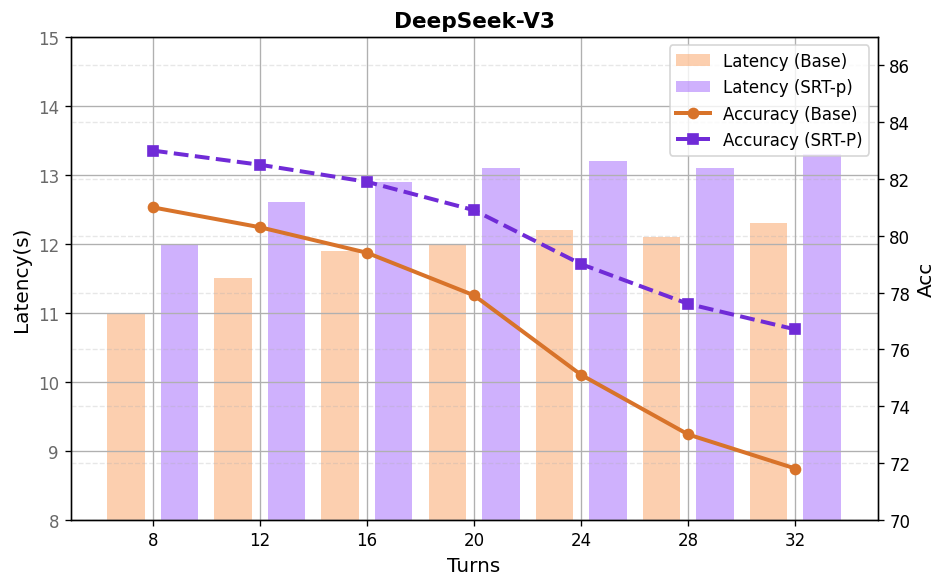}
    \caption{DeepSeek-V3}
\end{subfigure}
\hfill
\begin{subfigure}[t]{0.32\linewidth}
    \centering
    \includegraphics[width=\linewidth]{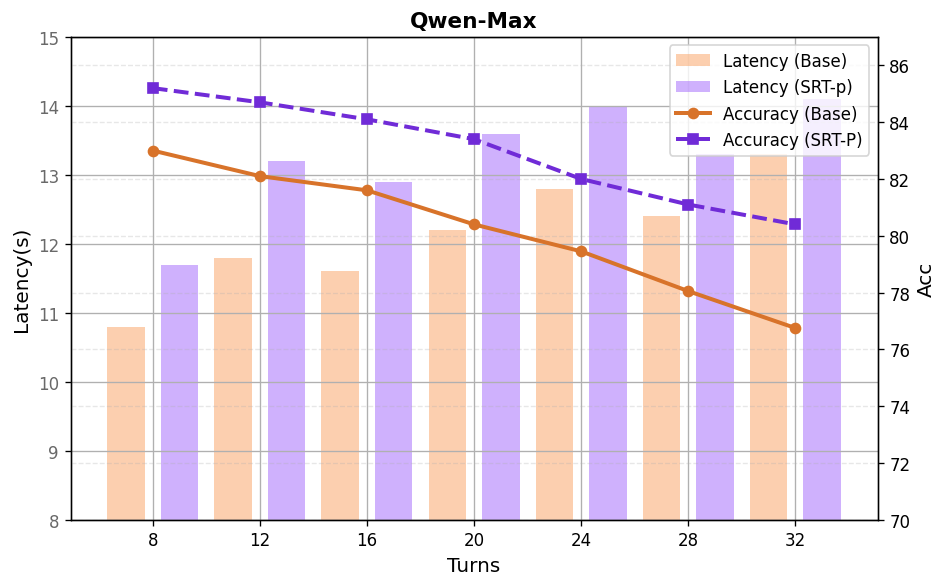}
    \caption{Qwen-Max}
\end{subfigure}
\hfill
\begin{subfigure}[t]{0.32\linewidth}
    \centering
    \includegraphics[width=\linewidth]{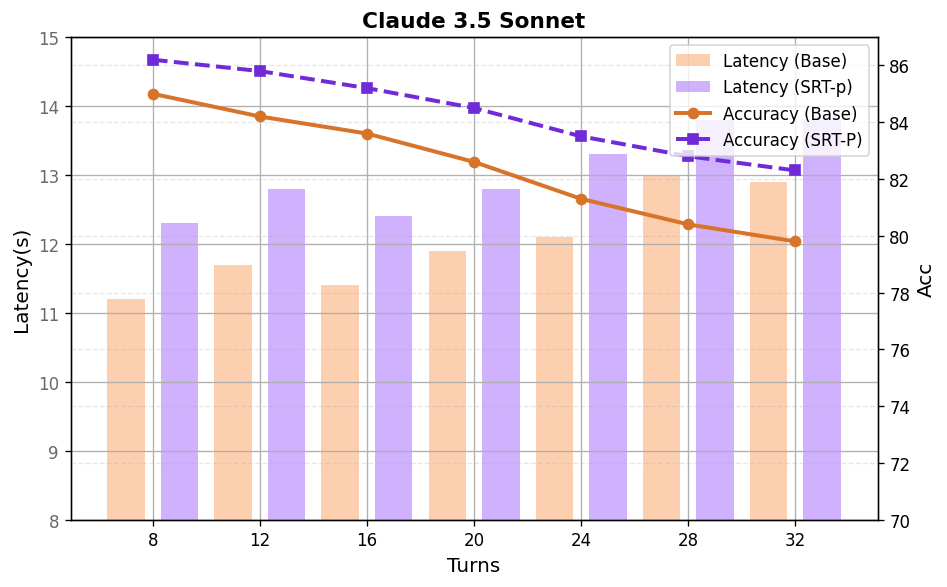}
    \caption{Claude 3.5 Sonnet}
\end{subfigure}
\caption{Efficiency of the SRT-P to closed-source LLMs.}
\label{lat}
\end{figure}
As shown in Fig. \ref{lat}, SRT-P achieves consistent accuracy improvements across all three closed-source LLms, with pronounced advantages in multi-turn buckets ($k \geq 24$) , where 2.7\% for DeepSeek-V3, 2.3\% for Qwen-Max, and 1.7\% for Claude 3.5 Sonnet improvement. The effectiveness stems from the $\texttt{<HIS>}$, which explicitly facilitates the selection and citation of dialogue-turn knowledge within black-box reasoning mechanisms, optimizing  attention allocation and consistency preservation, which shares consistent latent bias with the minimal sufficient history and correct answering objective of alignment in SRT.

\section{Conclusion}
In this paper, we introduced SRT, a novel framework designed to enhance the consistency and accuracy of multi-turn dialogue systems. This is achieved through a multi-stage process that combines fine-tuning to build foundational recall paradigm with alignment phase that refines reasoning using verifiable rewards. This results in an efficient and interpretable selection and utilization of historical knowledge mechanism that operates without external retrievers or indexes. Extensive experiments demonstrate that SRT achieves state-of-the-art performance across multiple dialogue benchmarks, offering a superior trade-off between accuracy and latency. Ablation studies further confirm that self-recall mechanism optimizes attention allocation and that reward strategy effectively guides the model's behavior. In dialogue agent applications, the current self-recall paradigm maintains multi-turn dialogue consistency well, while balancing compression costs and latency, and can be cost-effectively transferred to workflows.

\section{Acknowledgments}
This work was substantially supported by the National Natural Science Foundation of China (Grant No. U2336204, 62472075) and the Chengdu Industrial Chain Collaborative Innovation Project (Grant No. 2025-XT00-00017-GX).

\end{document}